\documentclass[9pt,conference]{IEEEtran}

\usepackage{cite}

\ifCLASSINFOpdf
  \usepackage[pdftex]{graphicx}
  \usepackage{epsfig}
  \usepackage{epstopdf}
  \usepackage{amssymb}

\else

\fi

\usepackage[cmex10]{amsmath}

\usepackage{algorithm2e}
\usepackage{algorithmicx}
\usepackage{bm}

\DeclareMathOperator*{\argmin}{arg\,min}

\newcommand{\mbr}[1]{\mathrm{\bm{\mathbf{#1}}}}

\begin{document}
%
\title{Image reconstruction from dense binary pixels}

\author{\IEEEauthorblockN{Or Litany$^*$}
\IEEEauthorblockA{orlitany@post.tau.ac.il}
\and
\IEEEauthorblockN{Tal Remez$^*$}
\IEEEauthorblockA{talremez@gmail.com}
\\
Tel-Aviv University\\
\tiny{$*$ Equal contributors}
\and
\IEEEauthorblockN{Alex Bronstein}
\IEEEauthorblockA{bron@eng.tau.ac.il}\\

}
\maketitle

\IEEEpeerreviewmaketitle


\section{Introduction}

The pursuit of smaller pixel sizes at ever increasing resolution in digital image sensors is mainly driven by the stringent price and form-factor requirements of sensors and optics in the cellular phone market. Recently, Eric Fossum proposed a novel concept of an image sensor with dense sub-diffraction limit one-bit pixels (\emph{jots}) \cite{fossum2005sub}, which can be considered a digital emulation of silver halide photographic film. This idea has been recently embodied as the EPFL Gigavision camera.
%


We denote by $\mbr{x}$ the radiant exposure at the camera aperture measured over a given time interval.
This exposure is subsequently degraded by the optical point spread function denoted by the operator $\mbr{H}$, producing the exposure of the sensor $\mbr{\lambda} = \mbr{Hx}$.
The number of photoelectrons $e_{jk}$ generated at pixel $j$ in time frame $k$ follows the Poisson distribution with the rate $\lambda_j$.
A binary pixel compares the accumulated charge against a pre-determined threshold $q_j$, outputting a one-bit measurement $b_{jk}$. Thus the probability of a single binary pixel $j$ to assume an "on" value in frame $k$ is $\mathrm{P}(b_{jk}=1)=\mathrm{P}(e_{jk} \geq {q_j} )$.
Our goal is to estimate an intensity field vector $\mbr{\hat{x}}$ best predicting $\mbr{x}$ given the measurement matrix $\mbr{B}$.

In \cite{EPFL-REPORT-166345}, a maximum likelihood (ML) approach was proposed. Assuming independent measurements, the negative likelihood function can be expressed as
\begin{equation}\label{likelihood}
\ell(\mbr{x} ; \mbr{B}) = \mathrm{const} - \sum_{kj} \log P(b_{jk} \mid q_j, \lambda_j),
\end{equation}
%
In ~\cite{EPFL-REPORT-166345} this objective is minimized w.r.t $\mbr{x}$ via standard iterative optimization techniques.

\section{Maximum Likelihood with Sparse Prior}

Since the ML approach assumes no prior, it needs a large amount of binary measurements in order to achieve good reconstruction.  Sparsity priors had been shown to give state of the art results in denoising tasks in general, and particularly in low light Poisson noise \cite{giryes2013sparsity,toEorNotToE}. In this work, we show that by introducing a similar sparsity spatial prior the number of measurements can be decreased significantly.
Assuming the light intensity $\mbr{\lambda}$ admits a non-linear sparse synthesis model $\mbr{\lambda} = \mbr{H}\rho(\mbr{Dz})$, with the dictionary $\mbr{D}$ and an element-wise non-linear transformation $\rho$ such as the non-negativity enforcing function from \cite{toEorNotToE}, we may construct the estimator as $\mbr{\hat{x}} = \rho(\mbr{D}\mbr{\hat{z}})$, where
\begin{equation}\label{problemFormulationWithSaprsity}
    \mbr{\hat{z}} = \argmin_{\mbr{z}}{\ell(\rho(\mbr{Dz}) ; \mbr{B}) + \mu \|\mbr{z}\|_1}.
\end{equation}
%
%
$\mu$ should be selected to best represent the tradeoff between the negative log-likelihood and the sparsity prior, in all experiments we selected $\mu$ empirically. The likelihood data fitting term is convex with a Lipschitz-continuous gradient (details are omitted due to lack of space), thus problem (\ref{problemFormulationWithSaprsity}) can be solved using proximal algorithms such as  FISTA \cite{beck2009fast}. Figures \ref{hdr} and \ref{Lena} show the significant improvement in image quality when using the sparse prior.

\section{Fast Approximation}

\begin{figure}[tb]
  \centering
  \includegraphics[width=1\columnwidth]{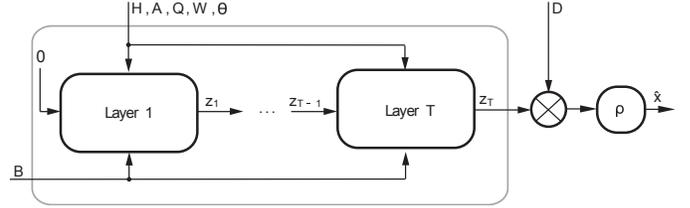}
  \caption{\small\textbf{MLNet architecture.}
A small number $T$ of ISTA iterations is unrolled into a feed-forward network. Each layer applies a non-linear transformation to the current iterate $\mbr{z}_t$, parametrized by $\mbr{H},\mbr{A},\mbr{Q},\mbr{W}$ and $\mbr{\theta}$. Training these parameters using standard backpropagation on a set of representative inputs allows the network to approximate the output of the underlying iterative algorithm with much lower complexity.  }
  \label{netProp}
\end{figure}

Iterative solutions of (\ref{problemFormulationWithSaprsity}) typically require hundreds of iterations to converge. This results in prohibitive complexity and unpredictable input-dependent latency unacceptable in real-time applications.
To overcome this limitation, we follow the approach advocated by \cite{gregor-icml-10} and \cite{sprechmann2012learning}, in which a small number of ISTA iterations are unrolled into a feed-forward neural network, that subsequently undergoes supervised training on typical inputs.
In our case, a single ISTA iteration can be written in the form

\begin{equation}
\mbr{z}_{t+1}\ = \sigma_\mbr{\theta}\left( \mbr{z}_{t} - \mbr{W}\rm{diag}(\rho'(\mbr{Q}\mbr{z}_{t}))\mbr{H}^\mathrm{T}\nabla_\mbr{\lambda} \ell (\rho(\mbr{A}\mbr{z}_{t}) ; \mbr{B} )\right),
\end{equation}
where $\mbr{A}=\mbr{Q}=\mbr{D}$, $\mbr{W}=\eta\mbr{D}^T$, $\mbr{\theta} = \mu\eta\mbr{1}$ ($\eta$ is the step size used by ISTA) and $\sigma_\mbr{\theta}$ is the two-sided shrinkage function.
Each such operation may be viewed as a single layer of the network parametrized by $\mbr{A},\mbr{Q},\mbr{W},\mbr{\theta}$, receiving $\mbr{z}_{t}$ as the input and producing $\mbr{z}_{t+1}$ as the output. Figure \ref{netProp} depicts the network architecture, henceforth referred to as MLNet.
When initializing the parameters as prescribed by the ISTA iteration and then adapting them by training that minimizes the reconstruction error of the entire network, the number of layers required to achieve comparable output quality on typical inputs is smaller by about two orders of magnitude than the number of corresponding ISTA iterations
(see Figure \ref{NN_complexity}). To the best of our knowledge, this is the first time a similar strategy is applied to reconstruction problems with a non-Euclidean data fitting term.

%

\section{Results}
\begin{figure}[]
	\label{hdr}
    \centering
        \begin{tabular}{ c c }
           \includegraphics[width = 0.22\textwidth]{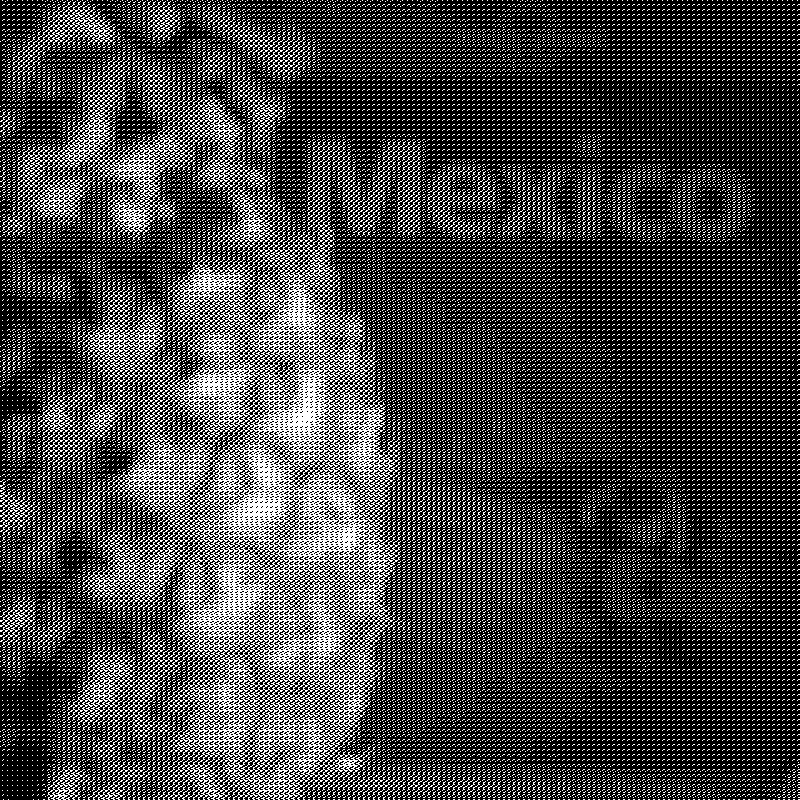} &
           \includegraphics[width = 0.22\textwidth]{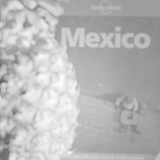} \\
           (a) Binary image & (b) Ground Truth \\
           \includegraphics[width = 0.22\textwidth]{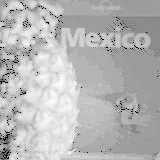} &
           \includegraphics[width = 0.22\textwidth]{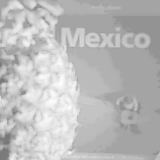} \\
           (c) ML (PSNR=$30.6$) & (d) FISTA (PSNR=$35.5$) \\
           \includegraphics[width = 0.22\textwidth]{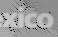} &
           \includegraphics[width = 0.22\textwidth]{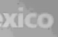} \\
           (e) ML zoom in & (f) FISTA zoom in \\
          \end{tabular}   \\
    \caption{\small \textbf{High dynamic range image reconstruction}. A five orders of magnitude HDR photoelectrons count image was assembled from multiple defocused raw images at different exposure times taken by a DSLR camera. (a) A single input binary image emulated by thresholding the photoelectrons count image using a predetermined threshold pattern, (b) Ground truth image produced by averaging and downsampling $10$ photoelectrons count images, (c) ML reconstruction without sparse prior (PSNR=$30.6$).  (d) ML reconstruction with a sparse prior (PSNR=$35.5$). (e) and (f) are zoom in versions of images (c) and (d) respectively. Images (b)-(f) are shown on a logarithmic scale.}


\end{figure}

\begin{figure}[t!]
\label{Lena}
\vspace{-40pt}
    \centering
        \begin{tabular}{ c c }
            \includegraphics[width = 0.22\textwidth]{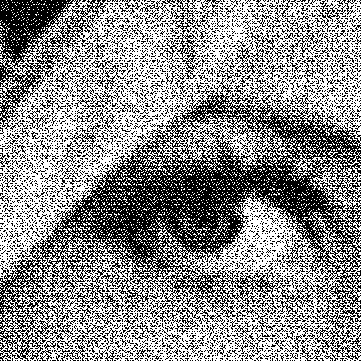} &
            \includegraphics[width = 0.22\textwidth]{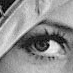} \\
            (a) Binary image & (b) Ground Truth \\
            \includegraphics[width = 0.22\textwidth]{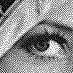} &
            \includegraphics[width = 0.22\textwidth]{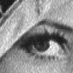} \\
            (c) ML (PSNR=$16.6$) & (d) MLNet (PSNR=$27.5$)\\
          \end{tabular}   \\ 
    \caption{\small \textbf{Low light reconstruction.}
        Lena's image was normalized to the range of $[0,10]$ from which four input binary images were simulated using a uniform threshold pattern with values $q_i\in\{1,...,10\}$.
        Depicted is a zoomed in fragment of the image: (a) input binary image, (b) low-resolution ground truth, (c) ML reconstruction (PSNR=$16.6$) and (d) MLNet reconstruction (PSNR=$27.5$). MLNet was trained on a disjoint set of patches from generic images.
}    
\end{figure}

\begin{figure}[]
\label{NN_complexity}
\includegraphics[width = 1\columnwidth]{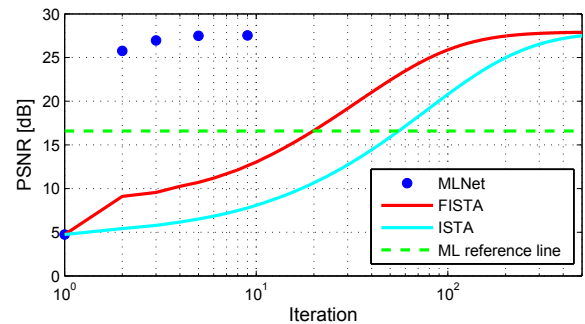}
\caption{\small \textbf{Bounded reconstruction latency comparison.} The plot shows the reconstruction quality for iterative algorithms (ISTA and FISTA) stopped after a given number of iterations, and the proposed MLNet with equivalent number of layers. As a reference, the performance of ML without the sparse prior is shown. Iteration $1$ represents the initialization for all algorithms. MLNet produces acceptable output quality and is about two orders of magnitude faster than ISTA and FISTA. The use of sparse prior has a clear advantage over pure ML. }
\end{figure}

Figure \ref{hdr} shows reconstruction results of an HDR image using ML with and without the sparse prior. FISTA was used to reconstruct overlapping $8\times 8$ patches that were subsequently averaged. An overcomplete dictionary was trained using $k$-SVD\cite{aharon2006img}.
Figure \ref{Lena} shows reconstruction results of an emulated low-light image. Figure \ref{NN_complexity} demonstrates the superiority of MLNet over iterative ML reconstruction on the same image. In all experiments $\hat{\mbr{x}}$ was initialized to the maximum dynamic range value.


\newpage

\bibliographystyle{IEEEtran}
\bibliography{binary_bib}

\end{document}